\documentclass{article}
\usepackage{spconf,amsmath,epsfig,amsfonts}
\usepackage{graphicx}
\usepackage{subcaption}
\let\OLDthebibliography\thebibliography
\renewcommand\thebibliography[1]{
  \OLDthebibliography{#1}
  \setlength{\parskip}{0pt}
  \setlength{\itemsep}{0pt plus 0.3ex}
}

\begin{document}\sloppy

\def\x{{\mathbf x}}
\def\L{{\cal L}}

\title{THOUSAND TO ONE: SEMANTIC PRIOR MODELING FOR CONCEPTUAL CODING}
%
\name{Jianhui Chang$^1$, Zhenghui Zhao$^2$, Lingbo Yang$^2$, Chuanmin Jia$^2$, Jian Zhang$^1$, Siwei Ma$^2$}
\address{$^1$ School of Electronic and Computer Engineering, Peking University, China 
\\
$^2$Institute of Digital Media, Peking University, China \\
\{jhchang, zhzhao, lingbo, cmjia, zhangjian.sz, swma\}@pku.edu.cn}

\maketitle

\begin{abstract}
Conceptual coding has been an emerging research topic recently, which encodes natural images into disentangled conceptual representations for compression.
However, the compression performance of the existing methods is still sub-optimal due to the lack of comprehensive consideration of rate constraint and reconstruction quality. 
To this end, we propose a novel end-to-end semantic prior modeling based conceptual coding scheme towards extremely low bitrate image compression, which leverages semantic-wise deep representations as a unified prior for entropy estimation and texture synthesis.
Specifically, we employ semantic segmentation maps as structural guidance for extracting deep semantic prior, which provides fine-grained texture distribution modeling for better detail construction and higher flexibility in subsequent high-level vision tasks. 
Moreover, a cross-channel entropy model is proposed to further exploit the inter-channel correlation of the spatially independent semantic prior, leading to more accurate entropy estimation for rate-constrained training.
The proposed scheme\footnote{For reproducible research, the source codes of our method will be made available when this paper is accepted.} achieves an ultra-high 1000$\times$ compression ratio, while still enjoying high visual reconstruction quality and versatility towards visual processing and analysis tasks.
\end{abstract}
\section{Introduction}
\label{sec:intro}
%
With the progress in deep generative models, conceptual coding~\cite{gregor2016towards,chang2019layered,chang2020conceptual} has emerged as a new paradigm for image compression beyond traditional signal-based image codecs, such as JPEG~\cite{pennebaker1992jpeg}, JPEG2000~\cite{rabbani2002jpeg2000}, HEVC~\cite{sze2014high} and other learning-based codecs~\cite{balle2018variational,minnen2018joint,balle2019end}.
%
%
Aiming at extracting decomposed conceptual representation from input visual data, conceptual coding not only achieves significant bitrate reduction over traditional codecs at comparable reconstruction quality, but also supports more flexible vision tasks.
Despite the achieved rapid progress, finding an efficient prior modeling and feature compression scheme under extreme conditions (\textit{e.g.}, 1000$\times$ compression ratio) still remain a considerable challenge.
%

More precisely, current image codecs typically involve entropy estimation through variational entropy-constrained training, where spatial dependencies in signal-based latent codes are often exploited for bitrate reduction~\cite{balle2018variational,minnen2018joint,lee2018context}.
%
As an approximation of bitrate, entropy can only be minimized properly if statistical dependencies over the compressed domain are properly captured.
However, the entropy modeling techniques are still under-explored for conceptual coding.
Specifically, existing methods~\cite{chang2019layered,chang2020conceptual} typically adopt a structure-texture dual-layered framework, yet the acquired conceptual codes are often compressed without effective rate optimization.
Furthermore, a single latent vector is usually leveraged to model global texture distribution of multiple semantic regions, where the intra-region similarity and cross-region independencies of texture codes are not fully exploited for entropy-constrained training.
In consequence, state-of-the-art conceptual coding schemes~\cite{chang2020conceptual} still exhibit inferior rate-distortion performance against current learning-based codecs (\textit{e.g.}, \cite{mentzer2020high}).
%
%
\begin{figure*}[t]
    \centering
    \vspace{-12mm}
    \includegraphics[width=1.0\linewidth]{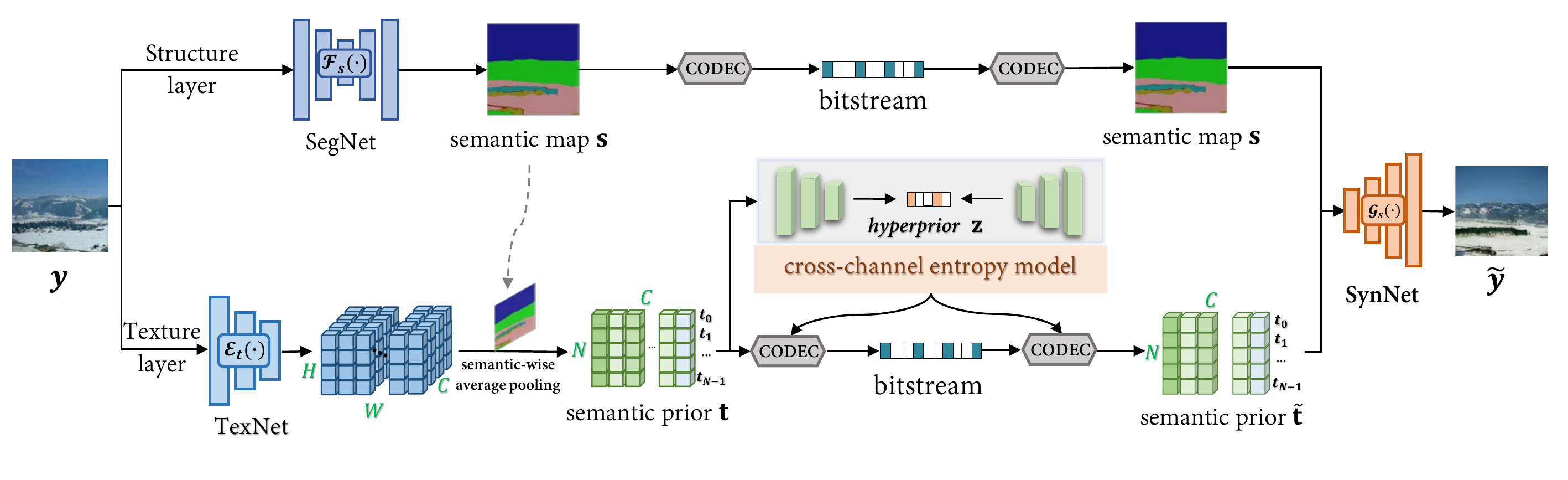}
    \vspace{-9mm}
    \caption{\textbf{Overview of our proposed semantic prior modeling based conceptual coding approach, consisting of a structure layer and a texture layer.} The structure layer is represented by semantic map, which is lossless coded and employed as guidance for extracting deep semantic prior and preserving structural information. The texture layer is modeled by semantic prior and its entropy is estimated by a cross-channel entropy model. The received semantic prior and map are integrated to synthesize the decoded image on the decoder side.}
    \label{fig:framework}
    \vspace{-5mm}
\end{figure*}
%

%
In this paper, we propose a novel semantic prior modeling based conceptual coding approach for ultra-low bitrate image compression by incorporating semantic-wise deep representations as a unified prior for both texture synthesis and entropy estimation.
As shown in Fig.~\ref{fig:framework}, instead of assuming a global texture code, we employ semantic segmentation maps as structural guidance to extract deep semantic prior within each individual semantic region.
On the one hand, the deep semantic prior models texture distributions semantic-wisely for finer texture representation and synthesis.
On the other hand, by taking advantage of semantic correlation, the conceptual representations are presented in a spatially independent form, which benefits more accurate entropy estimation.
%
%
%
%
%
Moreover, we propose a cross-channel entropy module with a hyperprior model to exploit inter-channel dependencies in semantic prior distribution for more accurate entropy estimation and higher bitrate reduction.
The proposed conceptual coding scheme is end-to-end trainable with entropy-constrained rate-distortion objectives, and is capable of achieving high reconstruction quality at extreme settings~(\emph{e.g.}, 1000$\times$ compression ratio). 
%
%
%
%
%
%
Our contributions can be summarized as follows:
\begin{itemize}
    \item To the best of our knowledge, we are the first to propose an end-to-end semantic prior modeling based conceptual coding scheme by extracting semantic-wise deep representations as a unified prior for both texture synthesis and entropy estimation, leading to significantly
    increased reconstruction quality and flexibility in content manipulation.
    \item We propose a cross-channel entropy model for effective hyperprior estimation and channel dependency reduction of semantic prior, allowing a more effective rate-distortion optimization with regard to entropy constraint.
    \item Extensive experiments demonstrate that the proposed method can achieve perceptually convincing reconstructions at extremely low bitrate (0.02-0.03 bpp, $\sim 1000\times$ compression ratio), as well as better support for various image analysis and manipulation tasks.
\end{itemize}
%
%

\section{Proposed Method}
%
\subsection{Semantic Prior Modeling}
In this paper, we adopt the structure-texture layered decomposition form to realize the conceptual coding framework.
Considering rate constraint and reconstruction quality comprehensively, we propose to model a semantic prior for texture representation and further entropy estimation.
In particular, as shown in Fig.~\ref{fig:framework}, input image $\mathbf{y}$ is processed into two basic visual features separately: 1) the structure layer characterized with the semantic segmentation map $\mathbf{s}$ which contains versatile information including structure layout, semantic category, location and shape, obtained by image segmentation networks ({SegNet}, \textit{e.g.}, PSPNet~\cite{zhao2017pyramid}) $\mathcal{F}_s(\cdot)$; and 2) the texture representations $\mathbf{t}$ modeled by semantic-wise deep prior extracted with a convolutional neural network (CNN) based feature extractor ({TexNet}, \textit{e.g.}, feature encoder in~\cite{wang2018high}) $\mathcal{E}_t(\cdot)$ with the guidance of semantic map $\mathbf{s}$. On the decoder side, the target image $\Tilde{\mathbf{y}}$ is reconstructed by integrating the decoded semantic prior $\mathbf{\Tilde{t}}$ and lossless semantic map $\mathbf{s}$ by (\textit{SynNet}, \textit{e.g.}, generator in~\cite{zhu2020sean}) $\mathcal{G}_s(\cdot)$, \textit{i.e.}, $\mathbf{\Tilde{y}}=\mathcal{G}_s(\mathbf{s,\Tilde{t}})$.
%

The process of extracting semantic prior is shown in Fig.~\ref{fig:framework}. A CNN-based feature extractor first transforms input images into intermediate features with the shape $C\times H\times W$, where $C,H,W$ correspond to channels, height and width, respectively. Then a semantic-wise average pooling layer is utilized to compute spatially average features under the guidance of semantic map, obtaining aggregated latent vectors corresponding to each semantic region as semantic prior. 
%
%
The shape of semantic prior is $C\times N$, where $N$ denotes the number of semantic class. The latent vectors $\left \{\mathbf{t}_0,\mathbf{t}_1,...,\mathbf{t}_{N-1}\right \}$ characterize the prior of semantic region $0,1,...,N-1$ correspondingly.
By taking advantage of semantic structure and average pooling, source latent feature maps reduce spatial dependencies and present as entropy modeling friendly semantic prior.
To further address internal channel dependencies in the latent vector of each semantic region, we propose a cross-channel entropy model and incorporate a hyperprior to model channel correlation for accurate entropy estimation as introduced in Sec.~\ref{entropy model}.
Combining reconstruction and entropy estimation tasks in training, our proposed semantic prior could effectively model texture distribution in an entropy modeling friendly form, which benefits both bitrate saving and reconstruction quality.
%
%
%
\subsection{Cross-channel Entropy Model}
\label{entropy model}
Plenty of entropy models have been introduced for joint rate-distortion optimization in learned image codecs.
%
Typically, Ball{\'e} \textit{et al.} developed a school of entropy models from a simple fully factorized model~\cite{balle2019end}, to conditional Gaussian mixture model incorporating hyperprior~\cite{balle2018variational} and context model~\cite{minnen2018joint}. The improvement of entropy model relies on the constantly further exploitation of dependencies in latent codes. 
%
%
However, different from latent codes obtained in signal-based nonlinear transform where spatial dependencies are mainly considered, conceptual representations demonstrate different correlation characteristics, urging the scheme of matching entropy model. 
%

\begin{figure}[t]
    \centering
    \vspace{0mm}
    \includegraphics[width=1.0\linewidth]{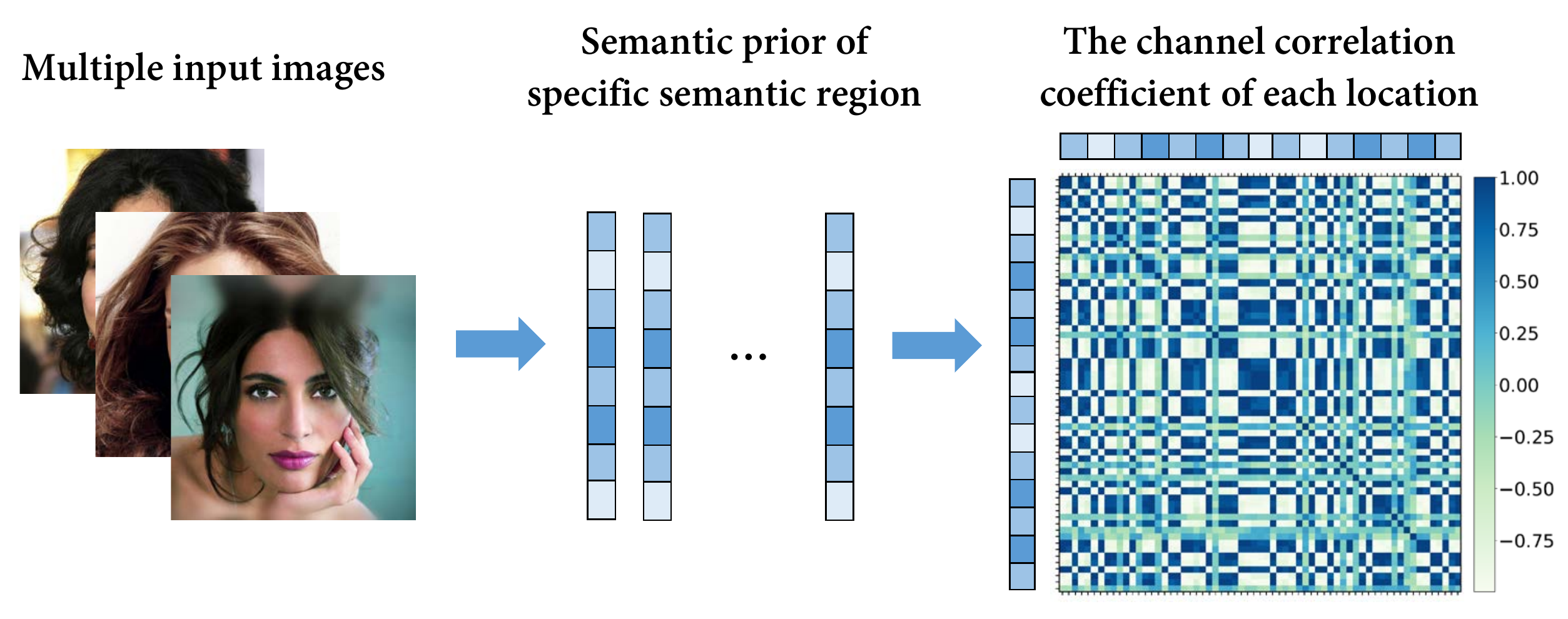}
    \vspace{-6mm}
    \caption{\textbf{The channel correlation coefficient.} Extracting semantic prior from a image set, the latent vectors of specific semantic region (\textit{e.g.}, hair) are separated to calculate the Pearson correlation coefficient~\cite{benesty2009pearson} among channels. Higher score (blue) shows higher positive correlation and lower score (white) shows higher negative correlation. The visualizations demonstrate high correlation among channels.} 
    \label{fig:correlation}
    \vspace{-5mm}
\end{figure}

Due to the responsibility for providing accurate semantic location guidance for reconstruction, semantic maps are lossless transmitted.
Thus, the proposed entropy model aims to model the probability distribution of semantic prior adaptively in training for the bit-saving purpose along with high reconstruction quality.
In essence, the learned semantic prior is presented in a spatial independent form by taking advantage of semantic correlation in the extraction process shown in Fig.~\ref{fig:framework}, leaving internal dependencies at channel dimension to further exploit. 
%
%
%
%
To quantitatively analyze the correlation, we extract the semantic prior from random $100$ images and separate the latent vectors of specific semantic region (\textit{e.g.}, hair) to calculate the Pearson correlation coefficient~\cite{benesty2009pearson} matrix as shown in Fig.~\ref{fig:correlation}, where the darker blue indicates the positive correspondence across channels in latent vectors and the example results demonstrate a high channel-wise correlation.
%
%
To this end, we propose a cross-channel entropy model, which incorporates a hyper-encoder to learn a \textit{cross-channel hyperprior} $\mathbf{z}$ to capture channel dependencies by three spatially invariant and channel-wise reduced convolutional layers, and a hyper-decoder to produce statistical parameters to conditional Gaussian mixture model for probability estimation.
As side information, the entropy of hyperprior is estimated by an independent density model as~\cite{balle2019end}.
%
%
By fully exploiting the statistical redundancy of deep semantic prior, the proposed cross-channel entropy model could effectively reduce the bitrate in training.
%
\subsection{Optimization Objectives}
%
In this paper, we introduce the rate-distortion optimization into conceptual coding~\cite{cover1999elements}. As shown in Fig.~\ref{fig:framework}, input image $\mathbf{y}$ is encoded into semantic maps $\mathbf{s}$ and semantic prior $\mathbf{t}$ respectively.
%
%
For entropy-constrained training, a cross-channel entropy model $\mathcal{P}_f(\cdot)$ is proposed to incorporate hyperprior $\mathbf{z}$ to estimate semantic prior entropy where the rate of quantized $\mathbf{\Tilde{z}}$ is estimated with factorized entropy model~\cite{balle2018variational} $\mathcal{P}_z(\cdot)$. 
The quantization is simulated with uniform noise as~\cite{balle2019end} in training and applies rounding algorithm directly in test.
The trainable rate constraint can be obtained as, 
\begin{equation}
    r(\mathbf{\Tilde{t}})=\mathbb{E}_{\mathbf{t}\sim p_T}\left \{-log_2(\mathcal{P}_f(\mathbf{\Tilde{t}}))\right \}+\mathbb{E}_{\mathbf{z}\sim p_Z}\left \{-log_2(\mathcal{P}_z(\mathbf{\Tilde{z}}))\right \}.
\end{equation}

On the decoder side, the decoded texture representation $\mathbf{\Tilde{t}}$ and lossless semantic map $\mathbf{s}$ are integrated by $\mathcal{G}_s(\cdot)$ to reconstruct the target image $\mathbf{\Tilde{y}}$, \textit{i.e.}, $\mathbf{\Tilde{y}}=\mathcal{G}_s(\mathbf{s,\Tilde{t}})$.
Since conceptual compression pursues appreciable visual reconstruction quality under extremely low bitrate rather than signal fidelity, and the pixel-wise similarity metrics prove to reduce signal distortion but impair perceptual quality~\cite{blau2019rethinking}, we employ the perceptual loss~\cite{johnson2016perceptual} $d_{P}$ and feature matching loss~\cite{wang2018high} $d_{FM}$ to form our distortion:
\begin{equation}
   d(\mathbf{y,\Tilde{y}})=\lambda_{P}d_{P}+\lambda_{FM}d_{FM}. 
\end{equation}
%

Furthermore, the conditional generative adversarial models (GANs~\cite{goodfellow2014generative}) are also employed to learn the distribution mapping from semantic map and semantic prior pair $\left \{ \mathbf{s,\Tilde{t}} \right \}$ to decoded image $\left \{ \mathbf{\Tilde{y}} \right \}$ under the condition $\mathbf{s}$, where the discriminator $\mathcal{D}$ is applied for the adversarial training. Additionally, the latent regression loss~\cite{chang2020conceptual} $\mathcal{L}_{r}$ is utilized as a regularization term to improve the semantic disentanglement of texture representations which can be verified with experiments. With parameterized models $\mathcal{E}_t,\mathcal{G}_s, \mathcal{D}, \mathcal{P}_f, \mathcal{P}_z$ and $\alpha, \beta$ as hyper-parameters for weight control, the loss objectives for rate-distortion and discriminator are shown as follows:
\begin{equation}
\label{RD}
    \mathfrak{L}_{\mathcal{E}_t,\mathcal{G}_s,\mathcal{P}_f}=\mathbb{E}_{y\sim p_{Y}}[\lambda r(\mathbf{\Tilde{t}})+d(\mathbf{y,\Tilde{y}})+\alpha \mathcal{L}_{r}-\beta log(\mathcal{D}(\mathbf{\Tilde{y},s}))],
\end{equation}
\begin{equation}
\label{discriminator}
    \mathfrak{L}_{\mathcal{D}}=\mathbb{E}_{y\sim p_{Y}}[-log(1-\mathcal{D}(\mathbf{\Tilde{y},s})+(-log(\mathcal{D}(\mathbf{y,s})))].
\end{equation}
\begin{figure*}[t]
    \centering
    \vspace{-13mm}
    \includegraphics[width=1.0\linewidth]{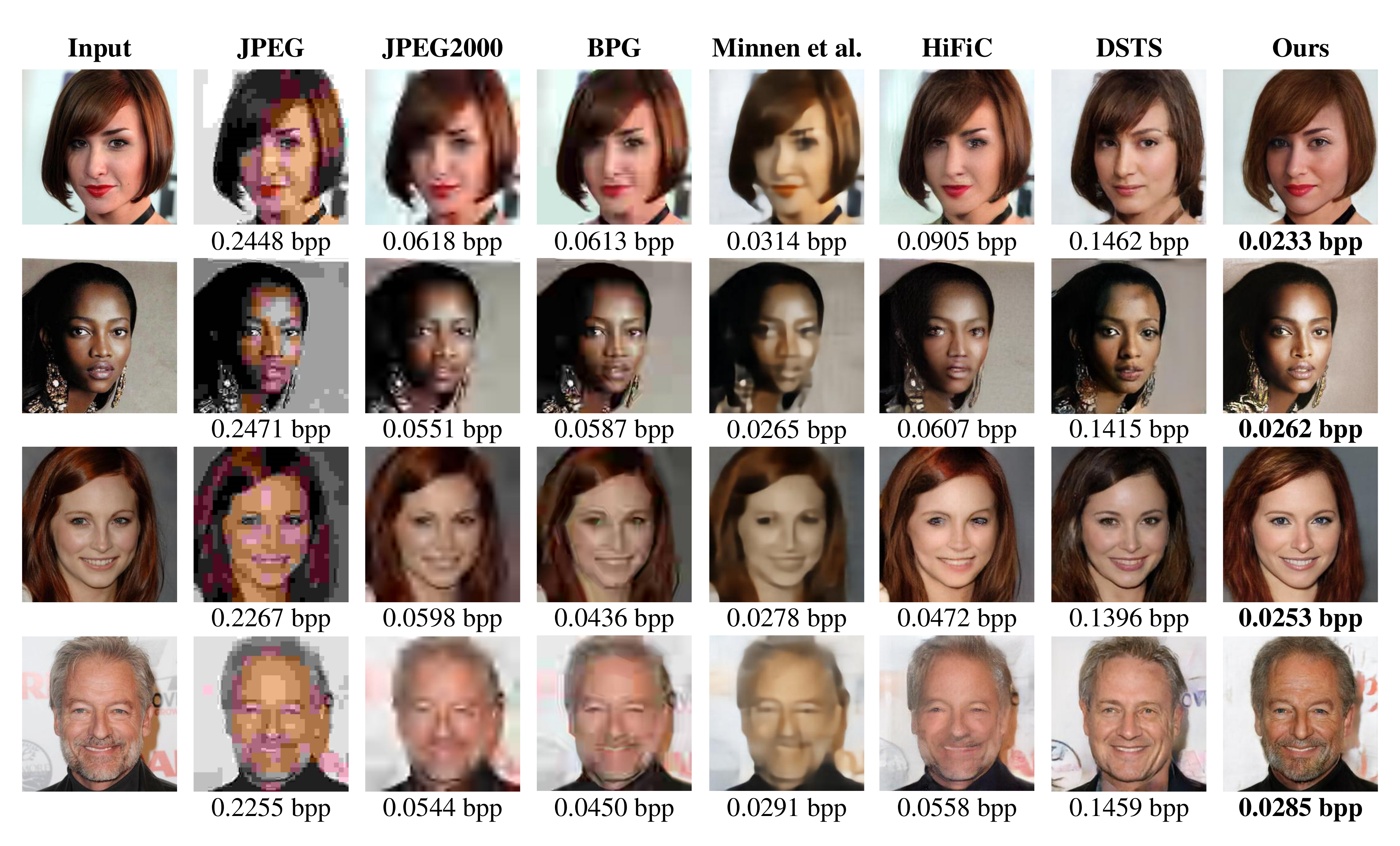}
    \vspace{-8mm}
    \caption{\textbf{Qualitative comparisons with baselines under low bitrate.} The \textbf{\textit{bpp}} of each image is reported under the corresponding image. The results show that our method outperforms others by achieving higher reconstruction quality with lower bitrate.}
    \label{fig:qualitative}
    \vspace{-4mm}
\end{figure*}

\section{Experiments}
\subsection{Experimental Settings}
\textbf{Networks.} The proposed hyper-encoder and hyper-decoder employ three $1\times 1$ convolutional layers respectively to learn the $16\times$ downscaled channel prior and corresponding mean and scale parameters. 
Besides, for the convenience of preliminary experiments, the \textit{TexNet}, \textit{SynNet} and discriminator are built upon~\cite{park2019semantic,zhu2020sean}. Note that we remove the Tanh activation and apply instance and spectral normalization in the decoder and discriminator.

\textbf{Dataset.} The proposed method is mainly evaluated on CelebAMask-HQ\footnote{https://github.com/switchablenorms/CelebAMask-HQ\label{celeba}} containing 19 semantic categories and 30,000 paired images of size $256\times 256$, with $24183$ as training set, $2824$ as testing set and $2993$ as validation set. Besides, ADE20K~\cite{zhou2017scene} is also utilized as additional dataset for discussion.

\textbf{Other settings.} 
The semantic map is lossless coded using FLIF\footnote{https://github.com/FLIF-hub/FLIF}. 
The channel dimension of texture representations is set to 64 and the quantization scale is set to 0.01 empirically for comparison.
We set the learning rate to $0.0001$ and the Adam optimizer~\cite{kingma2014adam} with default settings is used for training.
The parameters in Eq.~(\ref{RD}) are set as follows: $\alpha=1, \beta=1, \lambda_P=10.0, \lambda_{FM}=10.0$. 
The experiments are conducted on two NVIDIA Tesla V100 GPUs.
%
%
%
%
%
%
\subsection{Compression Performance Comparison}
\textbf{Baseline.}
We compare the compression performance with following typical approaches. 
For traditional codecs, widely used JPEG, JPEG2000 and HEVC-based BPG\footnote{https://bellard.org/bpg} are utilized for comparison. 
For exemplar learned image compression methods, we compare the proposed scheme with Minnen \textit{et al.}~\cite{minnen2018joint} and HiFiC~\cite{mentzer2020high} which are the state-of-the-art methods optimized without GANs and with GANs, respectively.
At last, our method is also compared to the state-of-the-art conceptual compression (named as DSTS)~\cite{chang2020conceptual} and model without cross-channel hyperprior as variants for ablation study.

\textbf{Qualitative results.}
The qualitative comparison results are shown in Fig.~\ref{fig:qualitative}.
Note that the bitrate of JPEG, BPG, Minnen \textit{et al.}~\cite{minnen2018joint} almost reach the lowest within the limit.
It can be seen that the proposed method achieves higher visual reconstruction quality and fidelity under extremely lower bitrate (average $0.0241$ bpp) compared to baselines.
In particular, compared with our models, the traditional codecs demonstrate severe degraded visual quality at higher bitrates (JPEG 9.9$\times$, JPEG2000 2.5$\times$, BPG 2.7$\times$).
%
%
Moreover, the decoded results from Minnen \textit{et al.}~\cite{minnen2018joint} show over-smoothing and severe distortion at similar bitrate.
%
Despite cooperated with adversarial training and LPIPS~\cite{zhang2018unreasonable} as perceptual distortion metric, at ultra-low bitrate range ($<$0.1 bpp), the reconstruction results appear apparent visual degradation and artifacts, leading to a less competitive model compared to ours.
%
%
%
\begin{figure}[t]
    \centering
    \vspace{0mm}
    \includegraphics[width=0.9\linewidth]{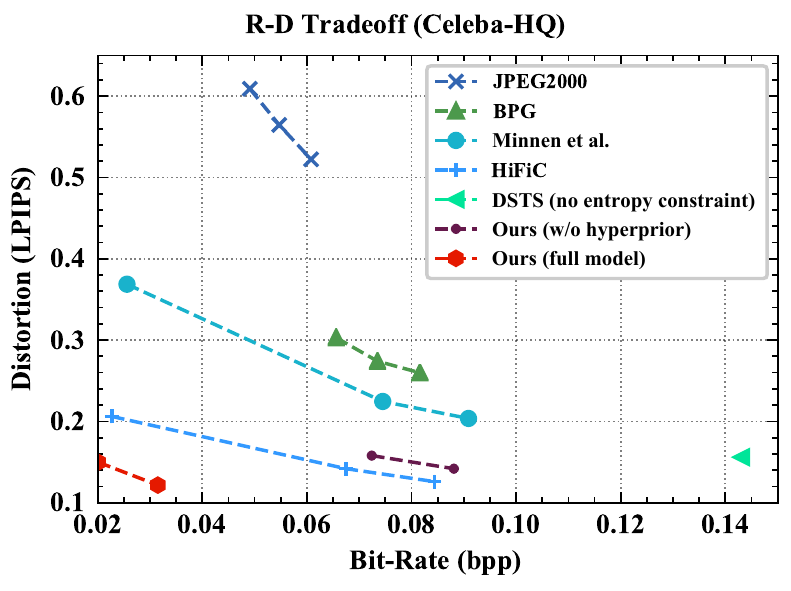}
    \vspace{-4mm}
    \caption{\textbf{Rate-distortion curves aggregated over the Celeba-HQ under extremely low bitrate range.} Lower LPIPS score demonstrates better visual quality.}
    \label{fig:quantitative}
    \vspace{-6mm}
\end{figure}
%

\textbf{Quantitative results.}
Fig.~\ref{fig:quantitative} shows RD curves over the publicly available Celeba-HQ\textsuperscript{\ref{celeba}} dataset by using LPIPS as the visual distortion metric at the ultra-low bitrate range ($<$0.1 bpp).
The rate-distortion (R-D) graphs compare our model to existing representative compression schemes. 
Even though using LPIPS as distortion loss brings a comparison advantage to HiFiC, the results clearly show our model achieves the best perceptual quality score of LPIPS while reaching an extremely low bitrate ever than before, outperforming other state-of-the-art methods.

\textbf{Ablation Study.} As the ablation study for the proposed model, we also show the RD performance of the model which replaces the proposed cross-channel entropy model with an independent Gaussian density model, and the model from DSTS~\cite{chang2020conceptual} without entropy constraint in Fig.~\ref{fig:quantitative}.
With fixed average bitrate $0.014$ bpp of lossless coded structure layer, the results show incorporating proposed cross-channel hyperprior into entropy model obtains an average bits-saving of $62.5\%$ over the non-hyperprior entropy model at similar visual quality, validating the effectiveness of the proposed cross-channel entropy model.
Due to lacking entropy constraint, the rate of DSTS is almost fixed at an average $0.1413$ bpp. For the texture layer, although the data volume for all semantic regions is $19$ times of it in DSTS, the actual bitrate for encoding them is only $4$ times than that in DSTS.
%

On the whole, our model achieves higher efficiency coding and better reconstruction by taking advantage of finer texture modeling and entropy-constrained training.

%
\begin{figure}[t]
    \centering
    \vspace{-10mm}
    \includegraphics[width=0.82\linewidth]{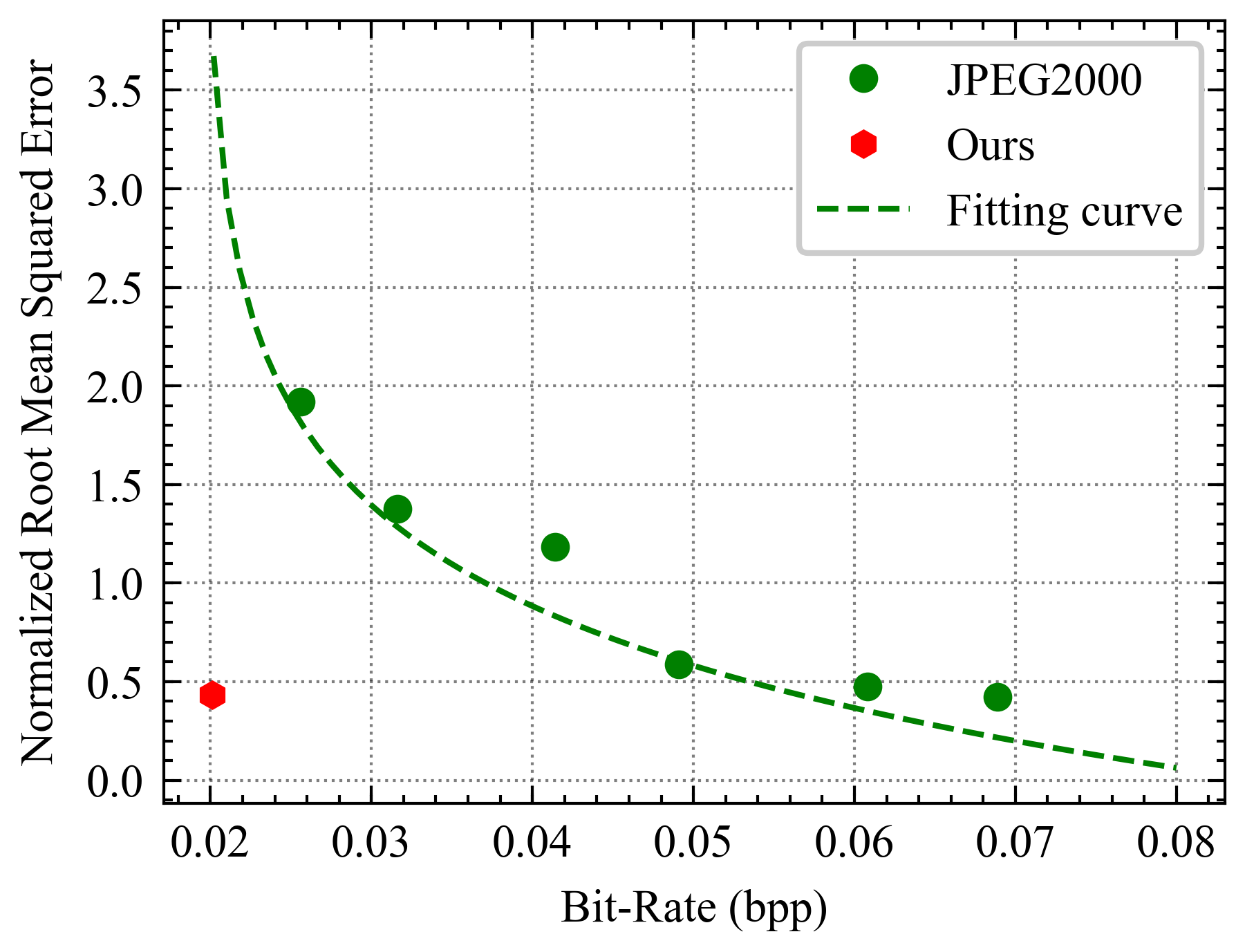}
    \vspace{-4mm}
    \caption{{The normalized root mean square error on facial landmark detection and bitrate of JPEG2000~\cite{rabbani2002jpeg2000} and proposed method.} Our method achieves $65.1\%$ bits saving at similar analysis accuracy.}
    \label{fig:detection}
    \vspace{-6mm}
\end{figure}
%


\subsection{Advantages for Vision Tasks}
\begin{figure}[t]
    \centering
    \vspace{-10mm}
    \includegraphics[width=1.0\linewidth]{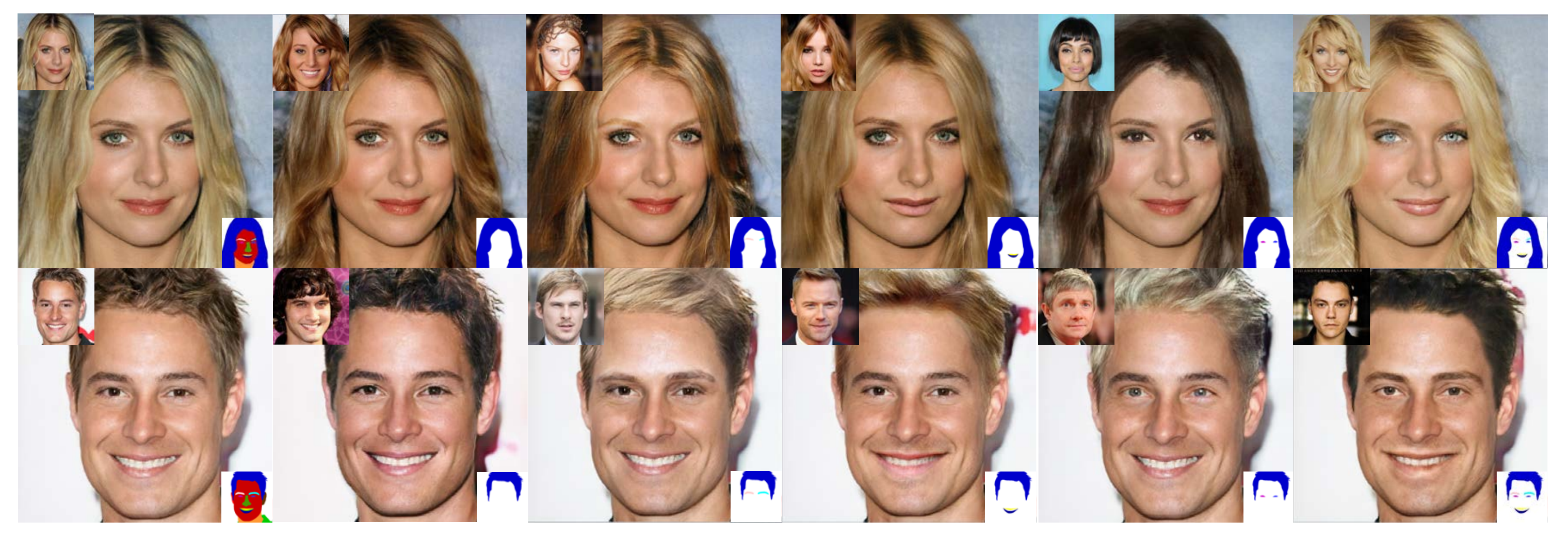}
    \vspace{-6mm}
    \caption{{The semantic-wise image manipulation results.} Reference images are shown at the top left corner and manipulated semantic regions are presented at the bottom right corner. The original synthesized images are shown in first column and manipulated results in last columns.}
    \label{fig:transfer}
    \vspace{-5mm}
\end{figure}

The advantages of proposed method in support of vision tasks can be presented in following two aspects.
On the one hand, under ``compression then analysis'' scenarios, our higher efficiency coding could benefit follow-up vision tasks performed on decoded images. 
For instance, we perform facial landmark detection on the decoded images from JPEG2000 and the proposed method and calculate the average normalized root mean squared error (NRMSE). 
As illustrated in Fig.~\ref{fig:detection}, our method outperforms JPEG2000 by achieving a lower NRMSE of $0.432$ under a lower bitrate of average $0.021$ bpp.
Particularly, our method can achieve $65.1\%$ bits saving at similar analysis accuracy, which demonstrates the superiority of the proposed method towards vision tasks.
On the other hand, benefited from the visual feature representations, conceptual coding has essential advantages over joint vision tasks in the compressed domain, corresponding to ``analysis then compression" scenarios.
In our approach, various visual features including structure, texture and semantic information can be applied to the analysis and content manipulation tasks directly without decoding, allowing higher efficiency and effectiveness.
%
%
Furthermore, compared to previous conceptual coding~\cite{chang2020conceptual}, besides providing direct semantic labels, our method can perform finer content manipulation with semantic prior as shown in Fig.~\ref{fig:transfer}.
\subsection{Generalization Discussion}
\begin{figure}[h]
    \centering
    \vspace{-3mm}
    \includegraphics[width=1.0\linewidth]{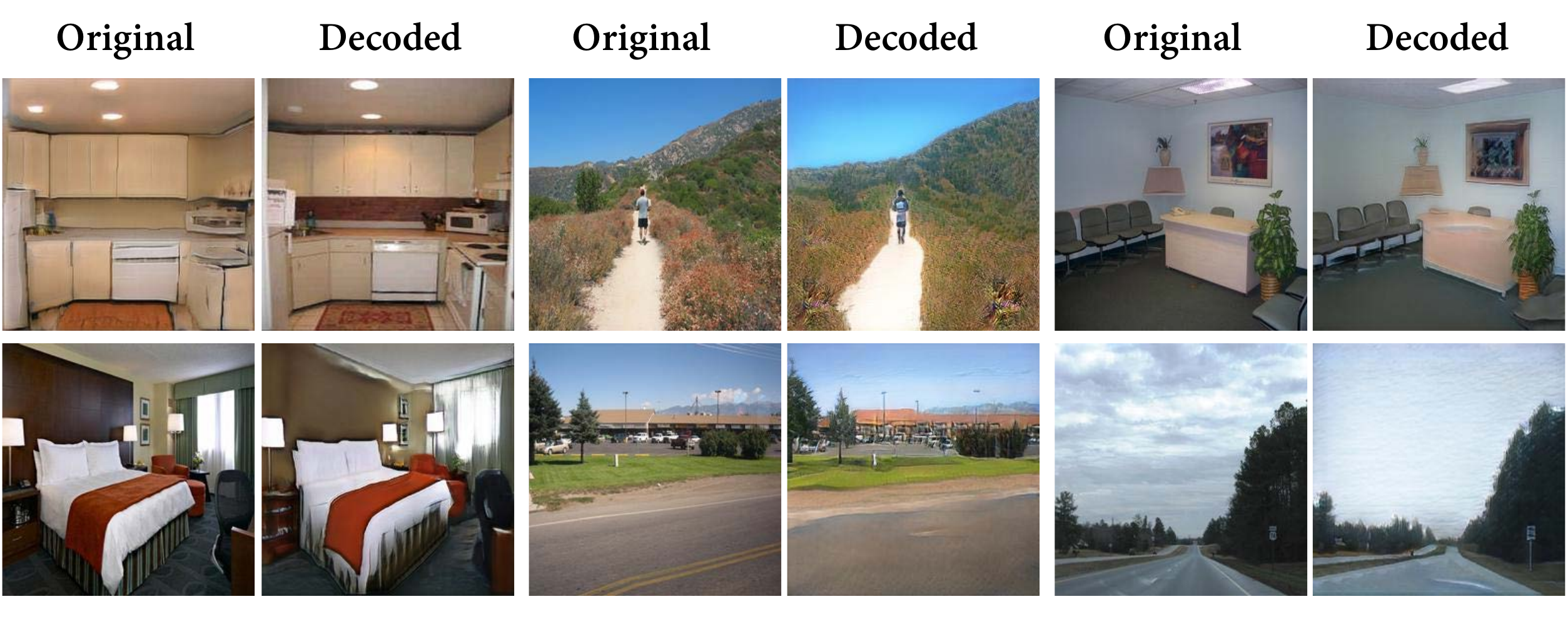}
    \vspace{-6mm}
    \caption{{The reconstruction cases of ADE20K dataset validate the generalization of proposed method.}}
    \label{fig:general}
    \vspace{-6mm}
\end{figure}

So far we have demonstrated outstanding compression performance and versatility of proposed method on facial dataset. 
Fig.~\ref{fig:general} shows the example cases of reconstruction results on ADE20K~\cite{zhou2017scene} dataset which consists of 150 semantic classes under the same training settings.
The scenarios in ADE20K contains much more complex texture and semantic information, validating the generalization and advantages of the proposed joint semantic prior conceptual coding model.
In essence, as a data-driven feature-based coding, our method can achieve better performance on domain-specific scenarios which appear structured visual characteristics.
%
%
\section{Conclusion}
This paper proposes a novel semantic prior modeling based conceptual coding approach which extracts semantic-wise deep representations to model texture distributions in an entropy modeling friendly form, to achieve ultra-low bitrate image compression with appreciable reconstruction quality.
Furthermore, we propose a cross-channel entropy model which exploits inter-channel correlation for accurate entropy estimation of semantic prior, leading to a high efficiency trainable model with rate-distortion optimization.
Qualitative and quantitative results demonstrate that the proposed scheme can perform extremely low bitrate image compression with high reconstruction quality and outperform the state-of-the-art methods.
The advantages of the proposed method over visual processing and understanding tasks are also analyzed and verified in our explorative experiments.
As a future direction, we would like to investigate more efficient and versatile algorithms for general scenes and video coding.
\bibliographystyle{IEEEbib}
\bibliography{main_v2}

\end{document}